\documentclass{article} 
\usepackage{nips13submit_e,times}
\usepackage{amsmath,amsfonts,amssymb}
\usepackage{amsfonts}
\usepackage[numbers]{natbib}
\usepackage{hyperref}
\newcommand{\Prob}[1]{\mathbb{P}\left( #1 \right)}
\newcommand{\Expect}[1]{\mathbf{E}\left[ #1 \right]}
\newcommand{\Expectwrt}[2]{\mathbf{E}_{#2}\left[ #1 \right]}
\newcommand{\Probwrt}[2]{\mathbb{P}_{#2}\left( #1 \right)}
\newcommand{\nrm}[1]{\left\| #1 \right\|}
\newcommand{\TV}[1]{\nrm{#1}_{\textrm{{\tiny \textup{TV}}}}}
\newtheorem{theorem}{Theorem}
\newtheorem{definition}{Definition}
\newtheorem{corollary}{Corollary}

\newcommand{\Extremals}{\mathcal{M}}
\newcommand{\Mixtures}{\mathcal{P}}
\newcommand{\MixingMeasure}{\pi}
\newcommand{\ExtremalProc}{\mu}
\newcommand{\MixedProc}{\rho}

\title{Predictive PAC Learning and Process Decompositions}

\author{Cosma Rohilla Shalizi\\
Statistics Department\\
Carnegie Mellon University\\
Pittsburgh, PA 15213 USA\\
\texttt{cshalizi@cmu.edu}\\
\And
Aryeh Kontorovich\\
Computer Science Department\\
Ben Gurion University\\
Beer Sheva 84105 Israel\\
\texttt{{karyeh@cs.bgu.ac.il}}
}

\newcommand{\fix}[1]{\marginpar{FIX}}

\nipsfinalcopy 

\begin{document}

\maketitle

\begin{abstract}
  We informally call a stochastic process learnable if it admits a
  generalization error approaching zero in probability for any concept class
  with finite VC-dimension (IID processes are the simplest example).  A mixture
  of learnable processes need not be learnable itself, and certainly its
  generalization error need not decay at the same rate. In this paper, we argue
  that it is natural in predictive PAC to condition not on the past
  observations but on the mixture component of the sample path. This definition
  not only matches what a realistic learner might demand, but also allows us to
  sidestep several otherwise grave problems in learning from dependent data. In
  particular, we give a novel PAC generalization bound for mixtures of
  learnable processes with a generalization error that is not worse than that
  of each mixture component.  We also provide a characterization of mixtures of
  absolutely regular ($\beta$-mixing) processes, of independent probability-theoretic interest.
\end{abstract}

\section{Introduction}

Statistical learning theory, especially the theory of ``probably approximately
correct'' (PAC) learning, has mostly developed under the assumption that data
are independent and identically distributed (IID) samples from a fixed, though
perhaps adversarially-chosen, distribution.  As recently as 1997,
\citet{Vidyasagar-learning} named extending learning theory to stochastic
processes of dependent variables as a major open problem.  Since then,
considerable progress has been made for specific classes of processes,
particularly strongly-mixing sequences and exchangeable sequences.  (Especially
relevant contributions, for our purposes, came from
\cite{Berti-Rigo-on-GC-for-exchangeable,Pestov-predictive-PAC} on
exchangeability, from
\cite{Yu-rates-of-convergence,Vidyasagar-on-learning-and-generalization} on
absolute regularity\footnote{Absolutely regular processes are ones where the
  joint distribution of past and future events approaches independence, in
  $L_1$, as the separation between events goes to infinity; see \S
  \ref{sec:mar} below for a precise statement and extensive discussion.
  Absolutely regular sequences are now more commonly called ``$\beta$-mixing'',
  but we use the older name to avoid confusion with the other sort of
  ``mixing''.}, and from
\cite{Adams-Nobel-VC-classes-under-ergodic,van-Handel-universal-GC} on
ergodicity; others include \cite{Modha-Masry-memory-universal,
  Meir-nonparametric-time-series, Karandikar-Vidyasagar-rates-of-UCEM,
  Gamarnik-PAC-for-Markov, SteinwartChristmann2009,
  Mohri-Rostamizdaeh-stability-bounds,
  Mohri-Rostamizadeh-rademacher-for-non-iid,
  Alquier-Wintenberger-weakly-dependent, London-Huang-Gettor-generalization,
  London-et-al-collective-stability}.)  Our goals in this paper are to point
out that many practically-important classes of stochastic processes possess a
special sort of structure, namely they are convex combinations of simpler,
extremal processes.  This both demands something of a re-orientation of the
goals of learning, and makes the new goals vastly easier to attain than they
might seem.

\paragraph{Main results}
Our main contribution is threefold: a conceptual definition of learning from
non-IID data (Definition \ref{def:pred}) and a technical result establishing
tight generalization bounds for mixtures of learnable processes (Theorem
\ref{thm:main}), with a direct corollary about exchangeable sequences
(Corollary \ref{cor:exchangeable}), and an application to mixtures of
absolutely regular sequences, for which we provide a new characterization.

\paragraph{Notation} $X_1, X_2, \ldots $ will be a sequence of dependent random
variables taking values in a common measurable space $\mathcal{X}$, which we
assume to be ``standard'' \cite[ch. 3]{Gray-ergodic-properties-2nd} to avoid
technicalities, implying that their $\sigma$-field has a countable generating
basis; the reader will lose little if they think of $\mathcal{X}$ as
$\mathbb{R}^d$.  (We believe our ideas apply to stochastic processes with
multidimensional index sets as well, but use sequences here.)  $X_i^j$ will
stand for the block $(X_i, X_{i+1}, \ldots X_{j-1}, X_j)$.  Generic
infinite-dimensional distributions, measures on $\mathcal{X}^{\infty}$, will be
$\ExtremalProc$ or $\MixedProc$; these are probability laws for $X_1^{\infty}$.
Any such stochastic process can be represented through the shift map $\tau:
\mathcal{X}^{\infty} \mapsto \mathcal{X}^{\infty}$ (which just drops the first
coordinate, $(\tau x)_t = x_{t+1}$), and a suitable distribution of initial
conditions.  When we speak of a set being invariant, we mean invariant under
the shift.  The collection of all such probability measures is itself a
measurable space, and a generic measure there will be $\MixingMeasure$.

\section{Process Decompositions}

Since the time of de Finetti and von Neumann, an important theme of the theory
of stochastic processes has been finding ways of representing complicated but
structured processes, obeying certain symmetries, as mixtures of simpler
processes with the same symmetries, as well as (typically) some sort of 0-1
law.  (See, for instance, the beautiful paper by
\citet{Dynkin-suff-stats-and-extreme-points}, and the statistically-motivated
\cite{Lauritzen-extreme-point-models}.)  In von Neumann's original ergodic
decomposition \cite[\S 7.9]{Gray-ergodic-properties-2nd}, stationary processes,
whose distributions are invariant over time, proved to be convex combinations
of stationary {\em ergodic} measures, ones where all invariant sets have either
probability 0 or probability 1.  In de Finetti's theorem \cite[ch.\
1]{Kallenberg-symmetries}, exchangeable sequences, which are invariant under
permutation, are expressed as mixtures of IID sequences\footnote{This is
  actually a special case of the ergodic decomposition \cite[pp.\
  25--26]{Kallenberg-symmetries}.}.  Similar results are now also known for
asymptotically mean stationary sequences \cite[\S
8.4]{Gray-ergodic-properties-2nd}, for partially-exchangeable sequences
\cite{Diaconis-Freedman-de-Finetti-for-Markov}, for stationary random fields,
and even for infinite exchangeable arrays (including networks and graphs)
\cite[ch.\ 7]{Kallenberg-symmetries}.

The common structure shared by these decompositions is as follows.
\begin{enumerate}
\item The probability law $\MixedProc$ of the composite but symmetric process
  is a convex combination of the simpler, extremal processes $\ExtremalProc \in
  \Extremals$ with the same symmetry.  The infinite-dimensional distribution of
  these extremal processes are, naturally, mutually singular.
\item Sample paths from the composite process are generated hierarchically,
  first by picking an extremal process $\ExtremalProc$ from $\Extremals$
  according to a some measure $\MixingMeasure$ supported on $\Extremals$, and
  then generating a sample path from $\ExtremalProc$.  Symbolically,
  \begin{eqnarray*}
    \ExtremalProc & \sim & \MixingMeasure\\
    X_1^{\infty} \mid \ExtremalProc & \sim & \ExtremalProc
  \end{eqnarray*}
\item Each realization of the composite process therefore gives information
  about only a single extremal process, as this is an invariant along each
  sample path.
\end{enumerate}

\section{Predictive PAC}

These points raise subtle but key issues for PAC learning theory.  Recall the
IID case: random variables $X_1, X_2, \ldots $ are all generated from a common
distribution $\ExtremalProc^{(1)}$, leading to an infinite-dimensional process
distribution $\ExtremalProc$.  Against this, we have a class $\mathcal{F}$ of
functions $f$.  The goal in PAC theory is to find a sample complexity
function\footnote{Standard PAC is defined as distribution-free, but here we
  maintain the dependence on $\mu$ for consistency with future notation.}
$s(\epsilon,\delta,\mathcal{F},\ExtremalProc)$ such that, whenever $n \geq s$,
\begin{equation}
  \Probwrt{\sup_{f\in\mathcal{F}}{\left|\frac{1}{n}\sum_{t=1}^{n}{f(X_t)} - \Expectwrt{f}{\ExtremalProc} \right|} \geq \epsilon}{\ExtremalProc} \leq \delta
\end{equation}
That is, PAC theory seeks finite-sample uniform law of large numbers for
$\mathcal{F}$.

Because of its importance, it will be convenient to abbreviate the supremum,
\[
\sup_{f\in\mathcal{F}}{\left|\frac{1}{n}\sum_{t=1}^{n}{f(X_t)} -
    \Expectwrt{f}{\ExtremalProc} \right|} \equiv \Gamma_n
\]
using the letter ``$\Gamma$'' as a reminder that when this goes to zero,
$\mathcal{F}$ is a Glivenko-Cantelli class (for $\ExtremalProc$).  $\Gamma_n$
is also a function of $\mathcal{F}$ and of $\ExtremalProc$, but we suppress
this in the notation for brevity.  We will also pass over the important and
intricate, but fundamentally technical, issue of establishing that $\Gamma_n$
is measurable (see \cite{Dudley-on-empirical-processes} for a thorough
treatment of this topic).

What one has in mind, of course, is that there is a space $\mathcal{H}$ of
predictive models (classifiers, regressions, \ldots) $h$, and that
$\mathcal{F}$ is the image of $\mathcal{H}$ through an appropriate loss
function $\ell$, i.e., each $f\in\mathcal{F}$ can be written as
\[
f(x) = \ell(x,h(x))
\]
for some $h \in \mathcal{H}$.  If $\Gamma_n \rightarrow 0$ in probability for
this ``loss function'' class, then empirical risk minimization is reliable.
That is, the function $\hat{f}_n$ which minimizes the empirical risk
$n^{-1}\sum_{t}{f(X_t)}$ has an expected risk in the future which is close to
the best attainable risk over all of $\mathcal{F}$,
$R(\mathcal{F},\ExtremalProc) =
\inf_{f\in\mathcal{F}}{\Expectwrt{f}{\ExtremalProc}}$.  Indeed, since when $n
\geq s$, with high ($\geq 1-\delta$) probability all functions have empirical
risks within $\epsilon$ of their true risks, with high probability the true
risk $\Expectwrt{\hat{f}_n}{\ExtremalProc}$ is within $2\epsilon$ of
$R(\mathcal{F},\ExtremalProc)$.  Although empirical risk minimization is not
the only conceivable learning strategy, it is, in a sense, a canonical one
(computational considerations aside). The latter is an immediate consequence of
the VC-dimension characterization of PAC learnability:
\begin{theorem}
  Suppose that the concept class $\mathcal{H}$ is PAC learnable from IID
  samples. Then $\mathcal{H}$ is learnable via empirical risk minimization.
\end{theorem}
\textsc{Proof:} Since $\mathcal{H}$ is PAC-learnable, it must necessarily have
a finite VC-dimension \cite{Blumer-et-al-learnability-and-VC}.  But for
finite-dimensional $\mathcal{H}$ and IID samples, $\Gamma_n\to0$ in probability
(see \cite{Boucheron-Bosquet-Lugosi} for a simple proof). This implies that the
empirical risk minimizer is a PAC learner for $\mathcal{H}$.  $\Box$

In extending these ideas to non-IID processes, a subtle issue arises,
concerning which expectation value we would like empirical means to converge
towards.  In the IID case, because $\ExtremalProc$ is simply the infinite
product of $\ExtremalProc^{(1)}$ and $f$ is a function on $\mathcal{X}$, we can
without trouble identify expectations under the two measures with each other,
and with expectations conditional on the first $n$ observations:
\[
\Expectwrt{f(X)}{\ExtremalProc} = \Expectwrt{f(X)}{\ExtremalProc^{(1)}} =
\Expectwrt{f(X_{n+1}) \mid X_1^n}{\ExtremalProc}
\]
Things are not so tidy when $\ExtremalProc$ is the law of a dependent
stochastic process.

In introducing a notion of ``predictive PAC learning'',
\citet{Pestov-predictive-PAC}, like \citet{Berti-Rigo-on-GC-for-exchangeable}
earlier, proposes that the target should be the conditional expectation, in our
notation $\Expectwrt{f(X_{n+1}) \mid X_1^n}{\ExtremalProc}$.  This however
presents two significant problems.  First, in general there is no single value
for this --- it truly is a function of the past $X_1^n$, or at least some part
of it.  (Consider even the case of a binary Markov chain.)  The other, and
related, problem with this idea of predictive PAC is that it presents learning
with a perpetually moving target.  Whether the function which minimizes the
empirical risk is going to do well by this criterion involves rather arbitrary
details of the process.  To truly pursue this approach, one would have to
actually learn the predictive dependence structure of the process, a quite
formidable undertaking, though perhaps not hopeless
\cite{Knight-predictive-view}.

Both of these complications are exacerbated when the process producing the data
is actually a mixture over simpler processes, as we have seen is very common in
interesting applied settings.  This is because, in addition to whatever
dependence may be present within each extremal process, $X_1^n$ gives (partial)
information about what that process is.  Finding $\Expectwrt{X_{n+1} \mid
  X_1^n}{\MixedProc}$ amounts to first finding all the individual conditional
expectations, $\Expectwrt{X_{n+1} \mid X_1^n}{\ExtremalProc}$, and then
averaging them with respect to the posterior distribution
$\MixingMeasure(\ExtremalProc \mid X_1^n)$.  This averaging over the posterior
produces additional dependence between past and future.  (See
\cite{Bialek-Nemenman-Tishby} on quantifying how much extra apparent Shannon
information this creates.)  As we show in \S \ref{sec:mar} below, continuous
mixtures of absolutely regular processes are far from being absolutely regular
themselves.  This makes it exceedingly hard, if not impossible, to use a single
sample path, no matter how long, to learn anything about global expectations.

These difficulties all simply dissolve if we change the target distribution.
What a learner should care about is not averaging over some hypothetical prior
distribution of inaccessible alternative dynamical systems, but rather what will
happen in the future of the particular realization which provided the training
data, and must continue to provide the testing data.  To get a sense of what
this is means, notice that for an ergodic $\ExtremalProc$,
\[
\Expectwrt{f}{\ExtremalProc} =
\lim_{m\rightarrow\infty}{\frac{1}{m}\sum_{i=1}^{m}{\Expect{f(X_{n+i}) \mid X_1^n}}}
\]
(from \cite[Cor.\ 4.4.1]{Lasota-Mackey}).  That is, matching expectations under
the process measure $\ExtremalProc$ means controlling the {\em long-run
  average} behavior, and not just the one-step-ahead expectation suggested in
\cite{Pestov-predictive-PAC,Berti-Rigo-on-GC-for-exchangeable}.  Empirical risk
minimization now makes sense: it is attempting to find a model which will work
well not just at the next step (which may be inherently unstable), but will
continue to work well, on average, indefinitely far into the future.

We are thus led to the following definition.
\begin{definition}
  \label{def:pred}
  Let $X_1^{\infty}$ be a stochastic process with law $\ExtremalProc$, and let
  $\mathcal{I}$ be the $\sigma$-field generated by all the invariant events.
  We say that $\ExtremalProc$ admits {\em predictive PAC learning} of a
  function class $\mathcal{F}$ when there exists a sample-complexity function
  $s(\epsilon,\delta,\mathcal{F},\ExtremalProc)$ such that, if $n \geq s$, then
  \[
  \Probwrt{\sup_{f\in\mathcal{F}}{\left| \frac{1}{n}\sum_{t=1}^{n}{f(X_t)} -
        \Expectwrt{f|\mathcal{I}}{\ExtremalProc}\right|} \geq
    \epsilon}{\ExtremalProc} \leq \delta
  \]
  A class of processes $\mathcal{P}$ admits of {\em distribution-free
    predictive PAC learning} if there exists a common sample-complexity
  function for all $\ExtremalProc \in \mathcal{P}$, in which case we write
  $s(\epsilon,\delta,\mathcal{F},\ExtremalProc) =
  s(\epsilon,\delta,\mathcal{F},\mathcal{P})$.
\end{definition}

As is well-known, distribution-free predictive PAC learning, in this sense, is
possible for IID processes ($\mathcal{F}$ must have finite VC dimension).  For
an ergodic $\ExtremalProc$, \cite{Adams-Nobel-VC-classes-under-ergodic} shows
that $s(\epsilon,\delta,\mathcal{F},\ExtremalProc)$ exist and is finite if and
only, once again, $\mathcal{F}$ has a finite VC dimension; this implies
predictive PAC learning, but not distribution-free predictive PAC.  Since
ergodic processes can converge arbitrarily slowly, some extra condition must be
imposed on $\mathcal{P}$ to ensure that dependence decays fast enough for each
$\ExtremalProc$.  A sufficient restriction is that all processes in
$\mathcal{P}$ be stationary and absolutely regular ($\beta$-mixing), with a common upper bound
on the $\beta$ dependence coefficients, as
\cite{Vidyasagar-on-learning-and-generalization,Mohri-Rostamizadeh-rademacher-for-non-iid}
show how to turn algorithms which are PAC on IID data into ones where are PAC
on such sequences, with a penalty in extra sample complexity depending on
$\ExtremalProc$ only through the rate of decay of correlations\footnote{We
  suspect that similar results could be derived for many of the weak dependence
  conditions of \cite{Weak-dependence}.}.  We may apply these familiar results
straightforwardly, because, when $\ExtremalProc$ is ergodic, all invariant sets
have either measure 0 or measure 1, conditioning on $\mathcal{I}$ has no
effect, and $\Expectwrt{f \mid \mathcal{I}}{\ExtremalProc} =
\Expectwrt{f}{\ExtremalProc}$.

Our central result is now almost obvious.

\begin{theorem}
\label{thm:main}
  Suppose that distribution-free prediction PAC learning is possible for a
  class of functions $\mathcal{F}$ and a class of processes $\Extremals$, with
  sample-complexity function $s(\epsilon,\delta,\mathcal{F},\mathcal{P})$.
  Then the class of processes $\Mixtures$ formed by taking convex mixtures
  from $\Extremals$ admits distribution-free PAC learning with the same sample
  complexity function.
\end{theorem}

\textsc{Proof:} Suppose that $n \geq s(\epsilon,\delta,\mathcal{F})$.  Then by
the law of total expectation,
\begin{eqnarray}
  \Probwrt{\Gamma_n \geq \epsilon}{\MixedProc} & = & \Expectwrt{\Probwrt{\Gamma_n \geq \epsilon \mid \ExtremalProc}{\MixedProc}}{\MixedProc}\\
  & = & \Expectwrt{\Probwrt{\Gamma_n \geq \epsilon}{\ExtremalProc}}{\MixedProc}\\
  & \leq & \Expectwrt{\delta}{\MixedProc} = \delta
\end{eqnarray}
$\Box$

In words, if the same bound holds for each component of the mixture, then it
still holds after averaging over the mixture.  It is important here that we are
only attempting to predict the long-run average risk along the continuation of
the same sample path as that which provided the training data; with this as our
goal, almost all sample paths looks like --- indeed, {\em are} --- realizations
of single components of the mixture, and so the bound for extremal processes
applies directly to them\footnote{After formulating this idea, we came across a
  remarkable paper by \citet{Wiener-prediction}, where he presents a
  qualitative version of highly similar considerations, using the ergodic
  decomposition to argue that a full dynamical model of the weather is neither
  necessary nor even helpful for meteorological forecasting. The same paper
  also lays out the idea of sensitive dependence on initial conditions, and the
  kernel trick of turning nonlinear problems into linear ones by projecting
  into infinite-dimensional feature spaces.}.  By contrast, there may be no
distribution-free bounds at all if one does not condition on $\mathcal{I}$.

A useful consequence of this innocent-looking result is that it lets us {\em
  immediately} apply PAC learning results for extremal processes to composite
processes, without any penalty in the sample complexity.  For instance, we have
the following corollary:
\begin{corollary}
  Let $\mathcal{F}$ have finite VC dimension, and have distribution-free sample
  complexity $s(\epsilon,\delta,\mathcal{F},\Extremals)$ for all IID measures
  $\ExtremalProc \in \mathcal{P}$.  Then the class $\mathcal{M}$ of
  exchangeable processes composed from $\mathcal{P}$ admit distribution-free
  PAC learning with the same sample complexity,
  \[
  s(\epsilon,\delta,\mathcal{F},\mathcal{P}) =
  s(\epsilon,\delta,\mathcal{F},\mathcal{M})
  \]
\label{cor:exchangeable}
\end{corollary}

This is in contrast with, for instance, the results obtained by
\cite{Berti-Rigo-on-GC-for-exchangeable,Pestov-predictive-PAC}, which both go
from IID sequences (laws in $\mathcal{P}$) to exchangeable ones (laws in
$\mathcal{M}$) at the cost of considerably increased sample complexity.  The
easiest point of comparison is with Theorem 4.2 of
\cite{Pestov-predictive-PAC}, which, in our notation, shows that
\[
s(\epsilon,\delta,\mathcal{M}) \leq s(\epsilon/2,\delta\epsilon,\mathcal{P})
\]
That we pay no such penalty in sample complexity because our target of learning
is $\Expectwrt{f \mid \mathcal{I}}{\ExtremalProc}$, not $\Expectwrt{f \mid
  X_1^n}{\MixedProc}$.  This means we do not have to use data points to narrow
the posterior distribution $\MixingMeasure(\ExtremalProc \mid X_1^n)$, and that
we can give a much more direct argument.

In \cite{Pestov-predictive-PAC}, Pestov asks whether ``one [can] maintain the
initial sample complexity'' when going from IID to exchangeable sequences; the
answer is yes, {\em if} one picks the right target.  This holds whenever the
data-generating process is a mixture of extremal processes for which learning
is possible.  A particularly important special case of this are the absolutely
regular processes.

\section{Mixtures of Absolutely Regular Processes}
\label{sec:mar}

Roughly speaking, an {\bf absolutely regular} process is one which is
asymptotically independent in a particular sense, where the joint distribution
between past and future events approaches, in $L_1$, the product of the
marginal distributions as the time-lag between past and future grows.  These
are particularly important for PAC learning, since much of the existing IID
learning theory translates directly to this setting.

To be precise, let $X_{-\infty}^{\infty}$ be a two-sided\footnote{We have
  worked with one-sided processes so far, but the devices for moving between
  the two representations are standard, and this definition is more easily
  stated in its two-sided version.} stationary process.  The $\beta$-dependence
coefficient at lag $k$ is
\begin{equation}
  \beta(k) \equiv \TV{P_{-\infty}^0 \otimes P_{k}^\infty - P_{-(1:k)}}
\end{equation}
where $P_{-(1:k)}$ is the joint distribution of $X_{-\infty}^0$ and
$X_{k}^{\infty}$, the total variation distance between the actual joint
distribution of the past and future, and the product of their marginals.
Equivalently \cite{Doukhan-on-mixing,Bradley-strong-mixing}
\begin{equation}
  \label{eq:altbeta}
  \beta(k) = \Expect{\sup_{B\in\sigma(X_k^{\infty})}{\Prob{B \mid X^0_{-\infty}} - \Prob{B}}}
\end{equation}
which, roughly, is the expected total variation distance between the marginal
distribution of the future and its distribution conditional on the past.

As is well known, $\beta(k)$ is non-increasing in $k$, and of course $\geq 0$,
so $\beta(k)$ must have a limit as $k \rightarrow\infty$; it will be convenient
to abbreviate this as $\beta(\infty)$.  When $\beta(\infty) = 0$, the process
is said to be beta mixing, or absolutely regular.  All absolutely regular
processes are also ergodic \cite{Bradley-strong-mixing}.

The importance of absolutely regular processes for learning comes essentially
from a result due to \citet{Yu-rates-of-convergence}.  Let $X_1^n$ be a part of
a sample path from an absolutely regular process $\mu$, whose dependence
coefficients are $\beta(k)$.  Fix integers $m$ and $a$ such that $2ma = n$, so
that the sequence is divided into $2m$ contiguous blocks of length $a$, and
define $\mu^{(m,a)}$ to be distribution of $m$ length-$a$ blocks.  (That is,
$\mu^{(m,a)}$ approximates $\mu$ by IID blocks.)  Then $|\mu(C) -
\mu^{(m,a)}(C)| \leq (m-1)\beta(a)$ \cite[Lemma 4.1]{Yu-rates-of-convergence}.
Since in particular the event $C$ could be taken to be $\left\{\Gamma_n >
  \epsilon\right\}$, this approximation result allows distribution-free
learning bounds for IID processes to translate directly into distribution-free
learning bounds for absolutely regular processes with bounded $\beta$
coefficients.

If $\mathcal{M}$ contains only absolutely regular processes, then a measure
$\MixingMeasure$ on $\mathcal{M}$ creates a $\MixedProc$ which is a mixture of
absolutely regular processes, or a MAR process.  It is easy to see that
absolute regularity of the component processes ($\beta_{\ExtremalProc}(k)
\rightarrow 0$) does not imply absolute regularity of the mixture processes
($\beta_{\MixedProc}(k) \not\rightarrow 0$).  To see this, suppose that
$\mathcal{M}$ consists of two processes $\ExtremalProc_0$, which puts unit
probability mass on the sequence of all zeros, and $\ExtremalProc_1$, which
puts unit probability on the sequence of all ones; and that $\MixingMeasure$
gives these two equal probability.  Then $\beta_{\ExtremalProc_i}(k) = 0$ for
both $i$, but the past and the future of $\MixedProc$ are not independent of
each other.

More generally, suppose $\MixingMeasure$ is supported on just two absolutely
regular processes, $\ExtremalProc$ and $\ExtremalProc^{\prime}$.  For each such
$\ExtremalProc$, there exists a set of typical sequences $T_{\ExtremalProc}
\subset \mathcal{X}^{\infty}$, in which the infinite sample path of
$\ExtremalProc$ lies almost surely\footnote{Since $\mathcal{X}$ is
  ``standard'', so is $\mathcal{X}^{\infty}$, and the latter's $\sigma$-field
  $\sigma(X_{-\infty}^{\infty})$ has a countable generating basis, say
  $\mathcal{B}$.  For each $B \in \mathcal{B}$, the set $T_{\ExtremalProc,B} =
  \left\{ x \in \mathcal{X}^{\infty} ~: ~
    \lim_{n\rightarrow\infty}{n^{-1}\sum_{t=0}^{n-1}{\mathbf{1}_{B}(\tau^t x)}}
    = \ExtremalProc(B)\right\}$ exists, is measurable, is dynamically
  invariant, and, by Bikrhoff's ergodic theorem,
  $\ExtremalProc(T_{\ExtremalProc,B}) = 1$ \cite[\S
  7.9]{Gray-ergodic-properties-2nd}.  Then $T_{\ExtremalProc} \equiv \bigcap_{B
    \in \mathcal{B}}{T_{\ExtremalProc,B}}$ also has $\ExtremalProc$-measure 1,
  because $\mathcal{B}$ is countable.}, and these sets do not
overlap\footnote{Since $\ExtremalProc \neq \ExtremalProc^{\prime}$, there
  exists at least one set $C$ with $\ExtremalProc(C) \neq
  \ExtremalProc^{\prime}(C)$.  The set $T_{\ExtremalProc,C}$ then cannot
  overlap at all with the set $T_{\ExtremalProc^{\prime},C}$, and so
  $T_{\ExtremalProc}$ cannot intersect $T_{\ExtremalProc^{\prime}}$.},
$T_{\ExtremalProc} \cap T_{\ExtremalProc^{\prime}} = \emptyset$.  This implies
that $\MixedProc(T_{\ExtremalProc}) = \MixingMeasure(\ExtremalProc)$, but
\[
\MixedProc(T_{\ExtremalProc} \mid X^{0}_{-\infty}) = \left\{
  \begin{array}{cc}
    1 & X^{0}_{-\infty} \in T_{\ExtremalProc}\\
    0 & \mathrm{otherwise}
  \end{array}
\right.
\]
Thus the change in probability of $T_{\ExtremalProc}$ due to conditioning on
the past is $\MixingMeasure(\ExtremalProc_1)$ if the selected component was
$\ExtremalProc_2$, and $1-\MixingMeasure(\ExtremalProc_1) =
\MixingMeasure(\ExtremalProc_2)$ if the selected component is
$\ExtremalProc_1$.  We can reason in parallel for $T_{\ExtremalProc_2}$, and so
the average change in probability is
\[
\MixingMeasure_1(1-\MixingMeasure_1) + \MixingMeasure_2(1-\MixingMeasure_2) =
2\MixingMeasure_1 (1-\MixingMeasure_1)
\]
and this must be $\beta_{\MixedProc}(\infty)$.  Similar reasoning when
$\MixingMeasure$ is supported on $q$ extremal processes shows
\[
\beta_{\MixedProc}(\infty) = \sum_{i=1}^{q}{\MixingMeasure_i (1-\MixingMeasure_i)}
\]
so that the general case is
\[
\beta_{\MixedProc}(\infty) = \int{[1 - \MixingMeasure(\{\ExtremalProc\})]d\MixingMeasure(\ExtremalProc)}
\]
This implies that if $\MixingMeasure$ has no atoms, $\beta_{\MixedProc}(\infty)
= 1$ always.  Since $\beta_{\MixedProc}(k)$ is non-increasing,
$\beta_{\MixedProc}(k) = 1$ for all $k$, for a continuous mixture of absolutely
regular processes.  Conceptually, this arises because of the use of
infinite-dimensional distributions for both past and future in the definition
of the $\beta$-dependence coefficient.  Having seen an infinite past is
sufficient, for an ergodic process, to identify the process, and of course the
future must be a continuation of this past.

MARs thus display a rather odd separation between the properties of individual
sample paths, which approach independence asymptotically in time, and the
ensemble-level behavior, where there is ineradicable dependence, and indeed
maximal dependence for continuous mixtures.  Any one realization of a MAR, no
matter how long, is indistinguishable from a realization of an absolutely
regular process which is a component of the mixture.  The distinction between a
MAR and a single absolutely regular process only becomes apparent at the level
of ensembles of paths.

It is desirable to characterize MARs.  They are stationary, but non-ergodic and
have non-vanishing $\beta(\infty)$.  However, this is not sufficient to
characterize them.  Bernoulli shifts are stationary and ergodic, but not
absolutely regular\footnote{They are, however, mixing in the sense of ergodic
  theory \cite{Lasota-Mackey}.}.  It follows that a mixture of Bernoulli shifts
will be stationary, and by the preceding argument will have positive
$\beta(\infty)$, but will not be a MAR.

Roughly speaking, given the infinite past of a MAR, events in the future become
asymptotically independent as the separation between them
increases\footnote{$\MixedProc$-almost-surely, $X^0_{-\infty}$ belongs to the
  typical set of a unique absolutely regular process, say $\mu$, and so the
  posterior concentrates on that $\mu$, $\pi(\cdot \mid
  X^{0}_{-\infty})=\delta_{\mu}$.  Hence $\rho(X_1^l,X_{l+k}^{\infty} \mid
  X^{0}_{-\infty}) = \mu(X^{l}_{-\infty},X_{l+k}^{\infty})$, which tends to
  $\mu(X^{l}_{-\infty})\mu(X_{l+k}^{\infty})$ as $k\rightarrow\infty$ because
  $\mu$ is absolutely regular.}.  A more precise statement needs to control
the approach to independence of the component processes in a MAR.  We say that
$\MixedProc$ is a {\bf $\tilde{\beta}$-uniform MAR} when $\MixedProc$ is a MAR,
and, for $\MixingMeasure$-almost-all $\ExtremalProc$, $\beta_{\ExtremalProc}(k)
\leq \tilde{\beta}(k)$, with $\tilde{\beta}(k) \rightarrow 0$.  Then if we
condition on finite histories $X^0_{-n}$ and let $n$ grow, widely separated
future events become asymptotically independent.

\begin{theorem}
  A stationary process $\MixedProc$ is a $\tilde{\beta}$-uniform MAR if and
  only if
  \begin{equation}
    \lim_{k\rightarrow\infty}{\lim_{n\rightarrow\infty}{\Expect{\sup_{l}{\sup_{B\in\sigma(X_{k+l}^{\infty})}{\MixedProc(B \mid X_1^l,X^0_{-n})
              - \MixedProc(B \mid X^0_{-n})}}}}} = 0
    \label{eqn:mar-criterion}
  \end{equation}
\end{theorem}
Before proceeding to the proof, it is worth remarking on the order of the
limits: for finite $n$, conditioning on $X^0_{-n}$ still gives a MAR, not a
single (albeit random) absolutely-regular process.  Hence the
$k\rightarrow\infty$ limit for fixed $n$ would always be positive, and indeed 1
for a continuous $\MixingMeasure$.

\textsc{Proof} ``Only if'': Re-write Eq.\ \ref{eqn:mar-criterion}, expressing
$\MixedProc$ in terms of $\MixingMeasure$ and the extremal processes:
\[
\lim_{k\rightarrow\infty}{
  \lim_{n\rightarrow\infty}{
    \Expect{
       \sup_{l}{
         \sup_{B\in\sigma(X_{k+l}^{\infty})}{
           \int{\left( \ExtremalProc(B \mid X_1^l,X^0_{-n}) - \ExtremalProc(B \mid X^0_{-n})\right)
            d\MixingMeasure(\ExtremalProc \mid X^0_{-n})
         }
       }
     }
   }
 }
}
\]
As $n\rightarrow\infty$, $\ExtremalProc(B \mid X^0_{-n}) \rightarrow
\ExtremalProc(B \mid X^0_{-\infty})$, and $\ExtremalProc(B \mid X_1^l,X^0_{-n})
\rightarrow \ExtremalProc(B \mid X^l_{-\infty})$.  But, in expectation, both of
these are within $\tilde{\beta}(k)$ of $\ExtremalProc(B)$, and so within
$2\tilde{\beta}(k)$.  ``If'': Consider the contrapositive.  If $\MixedProc$ is
not a uniform MAR, either it is a non-uniform MAR, or it is not a MAR at all.
If it is not a uniform MAR, then no matter what function $\tilde{\beta}(k)$
tending to zero we propose, the set of $\ExtremalProc$ with
$\beta_{\ExtremalProc} \geq \tilde{\beta}$ must have positive $\MixingMeasure$
measure, i.e., a positive-measure set of processes must converge arbitrarily
slowly.  Therefore there must exist a set $B$ (or a sequence of such sets)
witnessing this arbitrarily slow convergence, and hence the limit in Eq.\
\ref{eqn:mar-criterion} will be strictly positive.  If $\rho$ is not a MAR at
all, we know from the ergodic decomposition of stationary processes that it
must be a mixture of ergodic processes, and so it must give positive
$\MixingMeasure$ weight to processes which are not absolutely regular at all,
i.e., $\ExtremalProc$ for which $\beta_{\ExtremalProc}(\infty) > 0$.  The
witnessing events $B$ for these processes {\em a fortiori} drive the limit in
Eq.\ \ref{eqn:mar-criterion} above zero.  $\Box$

\section{Discussion and future work}

We have shown that with the right conditioning, a natural and useful notion of
predictive PAC emerges. This notion is natural in the sense that for a learner
sampling from a mixture of ergodic processes, the only thing that matters is
the future behavior of the component he is ``stuck'' in --- and certainly not
the process average over all the components. This insight enables us to adapt
the recent PAC bounds for mixing processes to mixtures of such processes.  An
interesting question then is to characterize those processes that are convex
mixtures of a given kind of ergodic process (de Finetti's theorem was the first
such characterization).  In this paper, we have addressed this question for
mixtures of uniformly absolutely regular processes. Another fascinating
question is how to extend predictive PAC to real-valued functions
\cite{Alon-et-al-scale-sensitive,Bartlett-Long-scale-sensitive-dimensions}.

\bibliographystyle{unsrtnat}
\bibliography{locusts}

\begin{thebibliography}{34}
\providecommand{\natexlab}[1]{#1}
\providecommand{\url}[1]{\texttt{#1}}
\expandafter\ifx\csname urlstyle\endcsname\relax
  \providecommand{\doi}[1]{doi: #1}\else
  \providecommand{\doi}{doi: \begingroup \urlstyle{rm}\Url}\fi

\bibitem[Vidyasagar(1997)]{Vidyasagar-learning}
Mathukumalli Vidyasagar.
\newblock \emph{A Theory of Learning and Generalization: With Applications to
  Neural Networks and Control Systems}.
\newblock Springer-Verlag, Berlin, 1997.

\bibitem[Berti and Rigo(1997)]{Berti-Rigo-on-GC-for-exchangeable}
Patrizia Berti and Pietro Rigo.
\newblock A {Glivenko}-{Cantelli} theorem for exchangeable random variables.
\newblock \emph{Statistics and Probability Letters}, 32:\penalty0 385--391,
  1997.

\bibitem[Pestov(2010)]{Pestov-predictive-PAC}
Vladimir Pestov.
\newblock Predictive {PAC} learnability: A paradigm for learning from
  exchangeable input data.
\newblock In \emph{Proceedings of the 2010 {IEEE} International Conference on
  Granular Computing ({GrC} 2010)}, pages 387--391, Los Alamitos, California,
  2010. {IEEE} Computer Society.
\newblock URL \url{http://arxiv.org/abs/1006.1129}.

\bibitem[Yu(1994)]{Yu-rates-of-convergence}
Bin Yu.
\newblock Rates of convergence for empirical processes of stationary mixing
  sequences.
\newblock \emph{Annals of Probability}, 22:\penalty0 94--116, 1994.
\newblock URL \url{http://projecteuclid.org/euclid.aop/1176988849}.

\bibitem[Vidyasagar(2003)]{Vidyasagar-on-learning-and-generalization}
M.~Vidyasagar.
\newblock \emph{Learning and Generalization: With Applications to Neural
  Networks}.
\newblock Springer-Verlag, Berlin, second edition, 2003.

\bibitem[Adams and Nobel(2010)]{Adams-Nobel-VC-classes-under-ergodic}
Terrence~M. Adams and Andrew~B. Nobel.
\newblock Uniform convergence of {Vapnik}-{Chervonenkis} classes under ergodic
  sampling.
\newblock \emph{Annals of Probability}, 38:\penalty0 1345--1367, 2010.
\newblock URL \url{http://arxiv.org/abs/1010.3162}.

\bibitem[van Handel(2013)]{van-Handel-universal-GC}
Ramon van Handel.
\newblock The universal {Glivenko}-{Cantelli} property.
\newblock \emph{Probability Theory and Related Fields}, 155:\penalty0 911--934,
  2013.
\newblock \doi{10.1007/s00440-012-0416-5}.
\newblock URL \url{http://arxiv.org/abs/1009.4434}.

\bibitem[Modha and Masry(1998)]{Modha-Masry-memory-universal}
Dharmendra~S. Modha and Elias Masry.
\newblock Memory-universal prediction of stationary random processes.
\newblock \emph{{IEEE} Transactions on Information Theory}, 44:\penalty0
  117--133, 1998.
\newblock \doi{10.1109/18.650998}.

\bibitem[Meir(2000)]{Meir-nonparametric-time-series}
Ron Meir.
\newblock Nonparametric time series prediction through adaptive model
  selection.
\newblock \emph{Machine Learning}, 39:\penalty0 5--34, 2000.
\newblock URL
  \url{http://www.ee.technion.ac.il/~rmeir/Publications/MeirTimeSeries00.pdf}.

\bibitem[Karandikar and Vidyasagar(2002)]{Karandikar-Vidyasagar-rates-of-UCEM}
Rajeeva~L. Karandikar and Mathukumalli Vidyasagar.
\newblock Rates of uniform convergence of empirical means with mixing
  processes.
\newblock \emph{Statistics and Probability Letters}, 58:\penalty0 297--307,
  2002.
\newblock \doi{10.1016/S0167-7152(02)00124-4}.

\bibitem[Gamarnik(2003)]{Gamarnik-PAC-for-Markov}
David Gamarnik.
\newblock Extension of the {PAC} framework to finite and countable {Markov}
  chains.
\newblock \emph{{IEEE} Transactions on Information Theory}, 49:\penalty0
  338--345, 2003.
\newblock \doi{10.1145/307400.307478}.

\bibitem[Steinwart and Christmann(2009)]{SteinwartChristmann2009}
Ingo Steinwart and Andreas Christmann.
\newblock Fast learning from non-i.i.d. observations.
\newblock In Y.~Bengio, D.~Schuurmans, John Lafferty, C.~K.~I. Williams, and
  A.~Culotta, editors, \emph{Advances in Neural Information Processing Systems
  22 [NIPS 2009]}, pages 1768--1776. MIT Press, Cambridge, Massachusetts, 2009.
\newblock URL \url{http://books.nips.cc/papers/files/nips22/NIPS2009_1061.pdf}.

\bibitem[Mohri and Rostamizadeh(2010)]{Mohri-Rostamizdaeh-stability-bounds}
Mehryar Mohri and Afshin Rostamizadeh.
\newblock Stability bounds for stationary $\phi$-mixing and $\beta$-mixing
  processes.
\newblock \emph{Journal of Machine Learning Research}, 11, 2010.
\newblock URL \url{http://www.jmlr.org/papers/v11/mohri10a.html}.

\bibitem[Mohri and
  Rostamizadeh(2009)]{Mohri-Rostamizadeh-rademacher-for-non-iid}
Mehryar Mohri and Afshin Rostamizadeh.
\newblock Rademacher complexity bounds for non-{I.I.D.} processes.
\newblock In Daphne Koller, D.~Schuurmans, Y.~Bengio, and L{\'e}on Bottou,
  editors, \emph{Advances in Neural Information Processing Systems 21 [NIPS
  2008]}, pages 1097--1104, 2009.
\newblock URL \url{http://books.nips.cc/papers/files/nips21/NIPS2008_0419.pdf}.

\bibitem[Alquier and Wintenberger(2012)]{Alquier-Wintenberger-weakly-dependent}
Pierre Alquier and Olivier Wintenberger.
\newblock Model selection for weakly dependent time series forecasting.
\newblock \emph{Bernoulli}, 18:\penalty0 883--913, 2012.
\newblock \doi{10.3150/11-BEJ359}.
\newblock URL \url{http://arxiv.org/abs/0902.2924}.

\bibitem[London et~al.(2012)London, Huang, and
  Getoor]{London-Huang-Gettor-generalization}
Ben London, Bert Huang, and Lise Getoor.
\newblock Improved generalization bounds for large-scale structured prediction.
\newblock In \emph{NIPS Workshop on Algorithmic and Statistical Approaches for
  Large Social Networks}, 2012.
\newblock URL
  \url{http://linqs.cs.umd.edu/basilic/web/Publications/2012/london:nips12asalsn/}.

\bibitem[London et~al.(2013)London, Huang, Taskar, and
  Getoor]{London-et-al-collective-stability}
Ben London, Bert Huang, Benjamin Taskar, and Lise Getoor.
\newblock Collective stability in structured prediction: Generalization from
  one example.
\newblock In Sanjoy Dasgupta and David McAllester, editors, \emph{Proceedings
  of the 30th International Conference on Machine Learning [ICML-13]},
  volume~28, pages 828--836, 2013.
\newblock URL \url{http://jmlr.org/proceedings/papers/v28/london13.html}.

\bibitem[Gray(2009)]{Gray-ergodic-properties-2nd}
Robert~M. Gray.
\newblock \emph{Probability, Random Processes, and Ergodic Properties}.
\newblock Springer-Verlag, New York, second edition, 2009.
\newblock URL \url{http://ee.stanford.edu/~gray/arp.html}.

\bibitem[Dynkin(1978)]{Dynkin-suff-stats-and-extreme-points}
E.~B. Dynkin.
\newblock Sufficient statistics and extreme points.
\newblock \emph{Annals of Probability}, 6:\penalty0 705--730, 1978.
\newblock URL \url{http://projecteuclid.org/euclid.aop/1176995424}.

\bibitem[Lauritzen(1984)]{Lauritzen-extreme-point-models}
Steffen~L. Lauritzen.
\newblock Extreme point models in statistics.
\newblock \emph{Scandinavian Journal of Statistics}, 11:\penalty0 65--91, 1984.
\newblock URL \url{http://www.jstor.org/pss/4615945}.
\newblock With discussion and response.

\bibitem[Kallenberg(2005)]{Kallenberg-symmetries}
Olav Kallenberg.
\newblock \emph{Probabilistic Symmetries and Invariance Principles}.
\newblock Springer-Verlag, New York, 2005.

\bibitem[Diaconis and Freedman(1980)]{Diaconis-Freedman-de-Finetti-for-Markov}
Persi Diaconis and David Freedman.
\newblock De {Finetti}'s theorem for {Markov} chains.
\newblock \emph{Annals of Probability}, 8:\penalty0 115--130, 1980.
\newblock \doi{10.1214/aop/1176994828}.
\newblock URL \url{http://projecteuclid.org/euclid.aop/1176994828}.

\bibitem[Dudley(1984)]{Dudley-on-empirical-processes}
R.~M. Dudley.
\newblock A course on empirical processes.
\newblock In \emph{{\'E}cole d'{\'e}t{\'e} de probabilit{\'e}s de
  {S}aint-{F}lour, {XII}--1982}, volume 1097 of \emph{Lecture Notes in
  Mathematics}, pages 1--142. Springer, Berlin, 1984.

\bibitem[Blumer et~al.(1989)Blumer, Ehrenfeucht, Haussler, and
  Warmuth]{Blumer-et-al-learnability-and-VC}
Anselm Blumer, Andrzej Ehrenfeucht, David Haussler, and Manfred~K. Warmuth.
\newblock Learnability and the {V}apnik-{C}hervonenkis dimension.
\newblock \emph{Journal of the Association for Computing Machinery},
  36:\penalty0 929--965, 1989.
\newblock \doi{10.1145/76359.76371}.
\newblock URL \url{http://users.soe.ucsc.edu/~manfred/pubs/J14.pdf}.

\bibitem[Boucheron et~al.(2005)Boucheron, Bousquet, and
  Lugosi]{Boucheron-Bosquet-Lugosi}
St{\'e}phane Boucheron, Olivier Bousquet, and G\'abor Lugosi.
\newblock Theory of classification: A survey of recent advances.
\newblock \emph{ESAIM: Probabability and Statistics}, 9:\penalty0 323--375,
  2005.
\newblock URL \url{http://www.numdam.org/item?id=PS_2005__9__323_0}.

\bibitem[Knight(1975)]{Knight-predictive-view}
Frank~B. Knight.
\newblock A predictive view of continuous time processes.
\newblock \emph{Annals of Probability}, 3:\penalty0 573--596, 1975.
\newblock URL \url{http://projecteuclid.org/euclid.aop/1176996302}.

\bibitem[Bialek et~al.(2001)Bialek, Nemenman, and
  Tishby]{Bialek-Nemenman-Tishby}
William Bialek, Ilya Nemenman, and Naftali Tishby.
\newblock Predictability, complexity and learning.
\newblock \emph{Neural Computation}, 13:\penalty0 2409--2463, 2001.
\newblock URL \url{http://arxiv.org/abs/physics/0007070}.

\bibitem[Lasota and Mackey(1994)]{Lasota-Mackey}
Andrzej Lasota and Michael~C. Mackey.
\newblock \emph{Chaos, Fractals, and Noise: Stochastic Aspects of Dynamics}.
\newblock Springer-Verlag, Berlin, 1994.
\newblock First edition, {\em Probabilistic Properties of Deterministic
  Systems}, Cambridge University Press, 1985.

\bibitem[Dedecker et~al.(2007)Dedecker, Doukhan, Lang, Le{\'o}n~R., Louhichi,
  and Prieur]{Weak-dependence}
J{\'e}r{\^o}me Dedecker, Paul Doukhan, Gabriel Lang, Jos{\'e}~Rafael
  Le{\'o}n~R., Sana Louhichi, and Cl{\'e}mentine Prieur.
\newblock \emph{Weak Dependence: With Examples and Applications}.
\newblock Springer, New York, 2007.

\bibitem[Wiener(1956)]{Wiener-prediction}
Norbert Wiener.
\newblock Nonlinear prediction and dynamics.
\newblock In Jerzy Neyman, editor, \emph{Proceedings of the Third {Berkeley}
  Symposium on Mathematical Statistics and Probability}, volume~3, pages
  247--252, Berkeley, 1956. University of California Press.
\newblock URL \url{http://projecteuclid.org/euclid.bsmsp/1200502197}.

\bibitem[Doukhan(1995)]{Doukhan-on-mixing}
Paul Doukhan.
\newblock \emph{Mixing: Properties and Examples}.
\newblock Springer-Verlag, New York, 1995.

\bibitem[Bradley(2005)]{Bradley-strong-mixing}
Richard~C. Bradley.
\newblock Basic properties of strong mixing conditions. {A} survey and some
  open questions.
\newblock \emph{Probability Surveys}, 2:\penalty0 107--144, 2005.
\newblock URL \url{http://arxiv.org/abs/math/0511078}.

\bibitem[Alon et~al.(1997)Alon, Ben-David, Cesa-Bianchi, and
  Haussler]{Alon-et-al-scale-sensitive}
Noga Alon, Shai Ben-David, Nicol{\`o} Cesa-Bianchi, and David Haussler.
\newblock Scale-sensitive dimensions, uniform convergence, and learnability.
\newblock \emph{Journal of the {ACM}}, 44:\penalty0 615--631, 1997.
\newblock \doi{10.1145/263867.263927}.
\newblock URL \url{http://tau.ac.il/~nogaa/PDFS/learn3.pdf}.

\bibitem[Bartlett and Long(1998)]{Bartlett-Long-scale-sensitive-dimensions}
Peter~L. Bartlett and Philip~M. Long.
\newblock Prediction, learning, uniform convergence, and scale-sensitive
  dimensions.
\newblock \emph{Journal of Computer and Systems Science}, 56:\penalty0
  174--190, 1998.
\newblock \doi{10.1006/jcss.1997.1557}.
\newblock URL \url{http://www.phillong.info/publications/more_theorems.pdf}.

\end{thebibliography}

\end{document}